\documentclass[runningheads]{llncs}

\usepackage{eccv}

\usepackage{eccvabbrv}

\usepackage{graphicx}
\usepackage{booktabs}

\usepackage[accsupp]{axessibility}  

\usepackage{bbding}  
\usepackage{multirow}
\usepackage{colortbl}
\usepackage{wrapfig}
\usepackage{threeparttable}
\usepackage{pifont}
\newcommand{\cmark}{\ding{51}} 
\newcommand{\xmark}{\ding{55}} 

\usepackage{orcidlink}

\begin{document}

\title{
Learning from Mistakes: 
Post-Training for Driving VLA with Takeover Data}

\titlerunning{TakeVLA: Post-training for driving VLA with takeover data}

\author{Yinfeng Gao\inst{1,2,3} \textsuperscript{$^\dagger$} 
\and
Deqing Liu\inst{3} \textsuperscript{$^\dagger$} \and
Qichao Zhang\inst{3} \textsuperscript{\Envelope} \and
Yupeng Zheng\inst{3} \and
Haochen Tian\inst{2,3} \and
Guang Li\inst{2} \and
Hangjun Ye\inst{2} \and
Long Chen\inst{2} \textsuperscript{$^\ddagger$} \and
Da-Wei Ding\inst{1} \textsuperscript{\Envelope} \and
Dongbin Zhao\inst{3}
}

\makeatletter
\def\@makefntext#1{\noindent #1}
\makeatother
\footnotetext{Work done while Yinfeng Gao interns at Xiaomi Embodied Intelligence Team.}

\authorrunning{Y.~Gao et al.}

\institute{University of Science and Technology Beijing \and
Xiaomi EV \and
Institute of Automation, Chinese Academy of Sciences
\\
\textsuperscript{$^\dagger$} Equal contribution.
\textsuperscript{$^\ddagger$} Project Leader.
\textsuperscript{\Envelope} Corresponding author.
}

\maketitle

\begin{figure}[!ht]
  \centering
  \includegraphics[height=6.3cm]{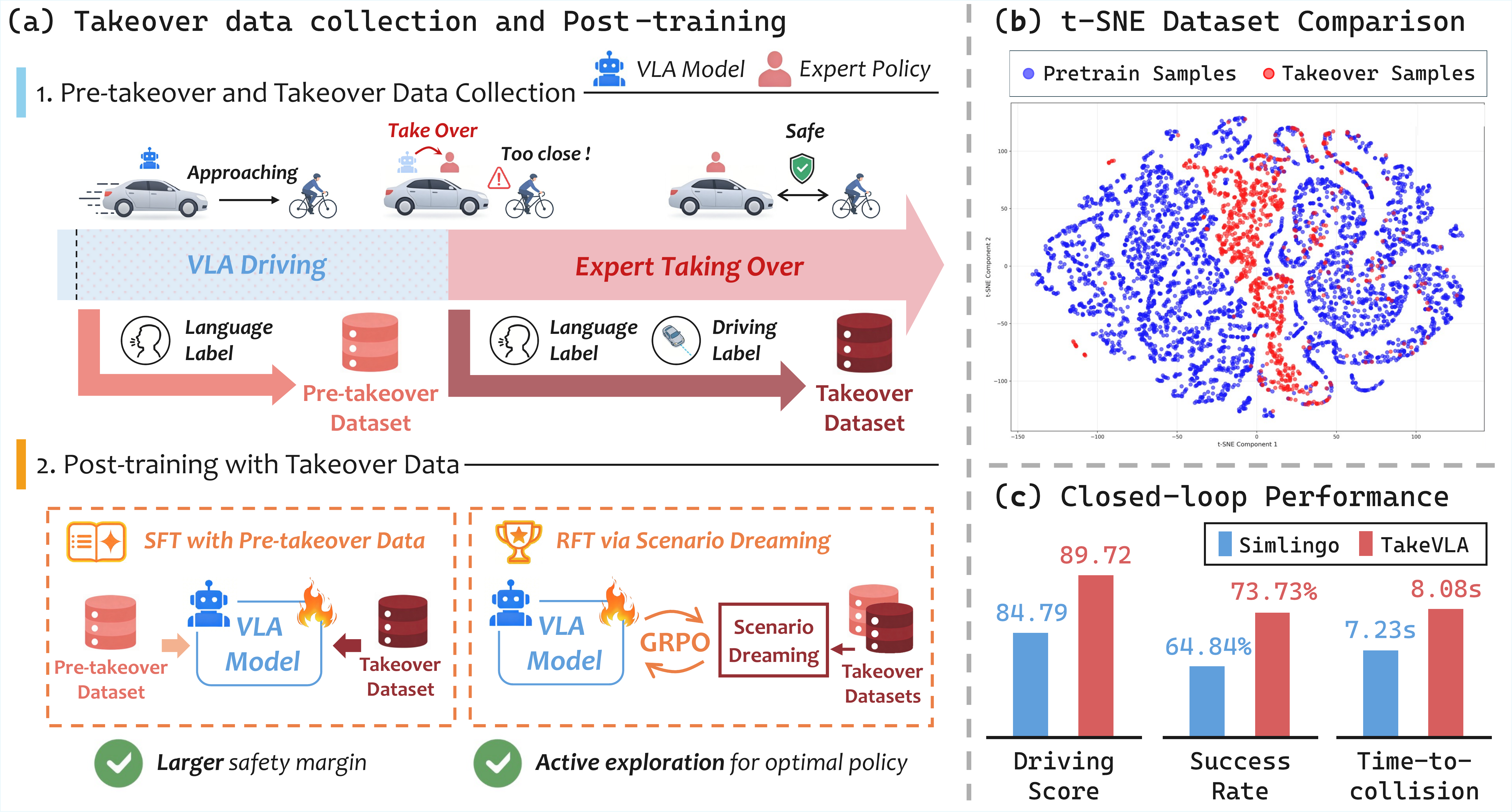}
  \caption{We propose \textbf{TakeVLA}, a post-training framework for driving VLA with takeover data.
  (a) TakeVLA iteratively collects takeover data via online expert interventions and conducts post-training on the collected data.
  It introduces \textit{pre-takeover} supervision for larger safety margins and \textit{Scenario Dreaming} for active exploration toward optimal policy.
  (b) t-SNE visualization shows that the takeover dataset (\textcolor{red}{red}) provides valuable OOD samples compared to the pretrain dataset (\textcolor{blue}{blue}).
  (c) Closed-loop evaluation on Bench2Drive demonstrates TakeVLA's superior performance, with a higher driving score, success rate, and significantly improved safety margins.
  }
  \label{fig:teaser}
\end{figure}

\begin{abstract}
Current Vision-Language-Action (VLA) paradigms in end-to-end autonomous driving rely on offline training from static datasets, leaving them vulnerable to distribution shift.
Recent post-training methods use takeover data to mitigate this by augmenting the dataset with high-quality expert takeover samples, yet they suffer from two key limitations: supervision restricted to the period after the takeover moments leads to policies with limited safety margins, and passive preference optimization lacks active exploration for optimal performance.
In this paper, we propose \textbf{TakeVLA}, a novel VLA post-training framework that overcomes these shortcomings through two complementary innovations.
First, we introduce pre-takeover language supervision, which allows the VLA to learn from mistakes proactively. 
By explicitly teaching the model about what to do in error-prone situations, we cultivate a precautionary mindset that anticipates hazards early and substantially enlarges safety margins.
Second, we propose Scenario Dreaming, a reinforcement fine-tuning paradigm that operates in reconstructed takeover scenarios, encouraging active exploration beyond mere preference fitting.
Experiments on the Bench2Drive benchmark demonstrate that TakeVLA achieves state-of-the-art closed-loop performance, surpassing the strong VLA baseline SimLingo by 4.93 in driving score, with an enhanced safety margin as evidenced by an 11.76\% increase in average TTC.

\end{abstract}

\section{Introduction}
\label{sec:intro}

Vision-Language-Action (VLA) models have emerged as a promising paradigm for end-to-end autonomous driving~\cite{chen2024end}. 
By aligning vision, language, and action spaces, VLAs offer more interpretable and generalizable decision-making than traditional modular pipelines or pure vision-action approaches~\cite{jiang2025alphadrive, zeng2025futuresightdrive}.

Despite these advantages, most VLA-based driving methods are trained on large-scale offline datasets via imitation learning (IL)~\cite{sima2024drivelm, zhou2025autovla, gao2019compare} or reinforcement learning (RL)~\cite{li2025recogdrive, rowe2025poutine, yang2025worldrft}, making them inherently vulnerable to distribution shift due to the lack of safety-critical or out-of-distribution (OOD) samples~\cite{tian2025simscale}.
Online RL-based methods attempt to address this through trial-and-error exploration in the environment~\cite{yang2025raw2drive, fu2025minddrive, gao2024piwm}, but this often leads to high safety risks and computational costs, as agents frequently enter dangerous or irrelevant regions.
Recently, post-training approaches leveraging expert takeover data have gained attention~\cite{cai2025predictive, liu2025takead, fang2025corevla}. 
These methods integrate expert interventions into the training loop: when the policy exhibits unsafe or suboptimal behavior, an expert driver takes over, providing high-quality corrective samples at low trial-and-error cost and effectively augmenting the dataset with valuable safety-critical and OOD demonstrations.
However, prior methods with takeover data suffer from two fundamental limitations.
First, supervision signals are restricted to the period after the takeover moments, yielding policies with insufficient safety margins.
Second, reliance on passive preference learning (\eg, DPO-style optimization \cite{rafailov2024dpo}) lacks active exploration, resulting in static fitting to demonstrated actions, which limits further performance gains.

We propose \textbf{TakeVLA}, a novel VLA post-training framework that addresses these gaps through two key innovations, as illustrated in \cref{fig:teaser}. 
First, to mitigate the limited safety margin issue in prior methods, we introduce \textbf{pre-takeover} supervision.
For each takeover segment, we extend labeling backward from the takeover moment to cover a preceding time window.
Since the expert has not yet intervened and no driving action supervision is available, we generate language supervision based on the specific takeover cause.
Supervised fine-tuning (SFT) on pre-takeover samples enables the model to anticipate hazards earlier through refined language outputs.
By leveraging the VLA model's inherent ability to align linguistic semantics with low-level actions~\cite{Shi-RSS-24}, the resulting language-conditioned actions substantially enlarge safety margins in error-prone situations.
Second, to overcome the lack of active exploration in preference-based optimization, we introduce a novel task termed \textbf{Scenario Dreaming}. 
By reconstructing takeover scenarios in a pseudo-simulation environment, we apply reinforcement fine-tuning (RFT) to encourage the exploration of superior language outputs, leading to more effective driving actions in safety-critical situations and going beyond mere fitting of expert preferences.

We train and evaluate TakeVLA on the challenging Bench2Drive \cite{jia2024bench2drive} closed-loop benchmark. 
Experimental results show consistent gains from both pre-takeover supervision and Scenario Dreaming. 
TakeVLA achieves a driving score of 89.72, representing a substantial improvement of 4.93 over the strong VLA baseline SimLingo \cite{renz2025simlingo}. 
Notably, without any rule-based post-processing, TakeVLA outperforms SimLingo by 7.25. 
Furthermore, pre-takeover training significantly enlarges safety margins, with average Time-to-Collision (TTC) increased by 11.76\% over SimLingo, validating the effectiveness of our designs.

Our main contributions are summarized as follows:
\begin{itemize}
\item We propose \textbf{TakeVLA}, a novel post-training framework for VLA-based autonomous driving. 
To our knowledge, it is the first method that exploits the inherent language-action alignment ability of VLAs to learn from pre-takeover contexts, enabling precautionary decision-making in language space and safe navigation in error-prone scenarios.
\item We introduce Scenario Dreaming, a novel reinforcement fine-tuning task that leverages reconstructed takeover scenarios for active exploration toward optimal policies, substantially enhancing performance in safety-critical scenarios.
\item Experiments on the Bench2Drive benchmark demonstrate that TakeVLA achieves state-of-the-art closed-loop performance, surpassing SimLingo by 4.93 in driving score, while substantially improving safety margins, yielding an 11.76\% increase in average TTC.
\end{itemize}

\section{Related Works}
\label{sec:related work}

\subsection{End-to-end Autonomous Driving}
End-to-end autonomous driving has seen rapid progress in recent years. 
Foundational works \cite{hu2023planning, jiang2023vad} established the paradigm of a holistic Transformer that maps raw sensor inputs directly to low-level driving actions, often augmented with auxiliary supervision for perception and prediction. 
Subsequent efforts have explored more efficient architectures \cite{sun2025sparsedrive, zheng2024genad}, multi-modal planning \cite{chen2024vadv2, liao2025diffusiondrive, wang2026meanfuser}, and self-supervised training \cite{li2024enhancing, zheng2025world4drive} to improve scalability and performance. 
However, these vision-only approaches lack rich world knowledge and explicit reasoning, resulting in limited interpretability and degraded performance in complex scenarios.
VLA models~\cite{kim2024openvla, BlackK-RSS-25} address this limitation by integrating pre-trained vision-language models to inject commonsense knowledge and semantic reasoning capabilities~\cite{zheng2026plan}. 
LMDrive \cite{shao2024lmdrive} pioneered language-guided closed-loop driving using natural-language instructions. 
ORION \cite{fu2025orion} introduced a generative trajectory decoder for more feasible actions. 
SimLingo \cite{renz2025simlingo} and Alpamayo-R1 \cite{wang2025alpamayo} aligned reasoning with actions through counterfactual language-action pairs and alignment-reward-based RL, respectively. 
AutoVLA \cite{zhou2025autovla} further enabled adaptive chain-of-thought reasoning conditioned on scene complexity via RL fine-tuning.
Nevertheless, these VLA methods rely exclusively on offline static datasets, leaving them vulnerable to distribution shift and poor recovery in unexpected situations.

\subsection{Post-training with Takeover Data}
Recent approaches leverage online expert interventions to collect high-quality takeover samples from safety-critical or unexpected situations without incurring real accidents, enabling low-risk policy improvement beyond static offline datasets.
PPL \cite{cai2025predictive} converts takeovers into predictive preference labels over future trajectories and applies contrastive preference optimization to boost policy performance. 
TakeAD \cite{liu2025takead} extends preference-based post-training to end-to-end policies through iterative refinement with takeover data, achieving significant gains in closed-loop performance. 
CoReVLA \cite{fang2025corevla} applies a similar preference-based optimization to VLA models to handle long-tail scenarios.
Despite these advances, existing methods share two key limitations: supervision restricted to the period after takeover moments yields policies with limited safety margins, while reliance on passive preference-based optimization lacks active exploration, resulting in static fitting to demonstrated actions.
Building on recent findings that language can effectively steer and refine low-level actions~\cite{Shi-RSS-24}, TakeVLA overcomes these limitations by performing pre-takeover supervision in the language space to enlarge safety margins, and by introducing Scenario Dreaming to enable active exploration for RL in reconstructed takeover scenarios.


\subsection{Reinforcement Fine-tuning for End-to-end Driving}
Reinforcement fine-tuning (RFT) enables pre-trained policies to discover optimal behaviors through active exploration~\cite{ouyang2022training, zhang2022trajgen, peng2024improving}. 
Existing RFT methods for end-to-end autonomous driving fall into two main categories. 
Offline approaches \cite{li2025recogdrive, zhou2025autovla, yang2025worldrft} train solely on static datasets and suffer from distribution shift due to the absence of real environment interactions. 
Online methods \cite{fu2025minddrive, yang2025raw2drive} rely on large-scale trial-and-error in real environments, incurring high safety risks and computational costs from exploring dangerous or irrelevant states.
In this work, we construct offline RL training scenarios from online-collected expert takeover data. 
These takeovers naturally constrain exploration costs by preventing actual accidents, while providing scene information to reconstruct safety-critical takeover scenarios for safe trial-and-error during reinforcement fine-tuning.

\section{Problem Formulation}
\label{sec:problem}
We formulate VLA-based end-to-end autonomous driving as a Partially Observable Markov Decision Process (POMDP)~\cite{sutton1998reinforcement}, defined by the tuple $\langle \mathcal{O}, \mathcal{S}, \mathcal{A}, P, O,$ $r,\gamma \rangle$.
Here, $\mathcal{O}$, $\mathcal{S}$, and $\mathcal{A}$ denote the observation, state, and action spaces, respectively.
$P$ and $O$ are the state transition and observation functions, $r$ is the reward function, and $\gamma \in [0,1)$ is the discount factor.
Concretely, the observation space is $\mathcal{O} = \{ o_{l}, o_{v}\}$, comprising natural language instructions $o_{l}$ and visual inputs $o_{v}$. 
The state space $\mathcal{S}$ represents the unobservable ground-truth environmental state.
The action space is $\mathcal{A} = \{a_{l}, a_{d}\}$, where $a_{l}$ is the generated language output consisting of a meta-action and an accompanying reason, and $a_{d}$ denotes the physical driving 
action.

The learned VLA policy is parameterized as $ \pi(a_{l}, a_{d} \mid o_{l}, o_{v})$.
In practice, expert takeover policies are typically human-provided. 
To ensure behavioral consistency, optimality, and scalability, we adopt the privileged rule-based \textit{PDM-Lite} \cite{sima2024drivelm, zimmerlin2024hidden} expert policy $\hat{\pi}(a_d \mid s), s \in \mathcal{S}$, which generates expert driving actions $\hat{a}_d$ using full simulator ground-truth information.
During online interaction, the predicted driving actions $a_{d}$ are converted via a low-level PID controller into throttle, brake, and steering commands applied to the ego-vehicle.

\begin{figure}[tb]
  \centering
  \includegraphics[height=4.5cm]{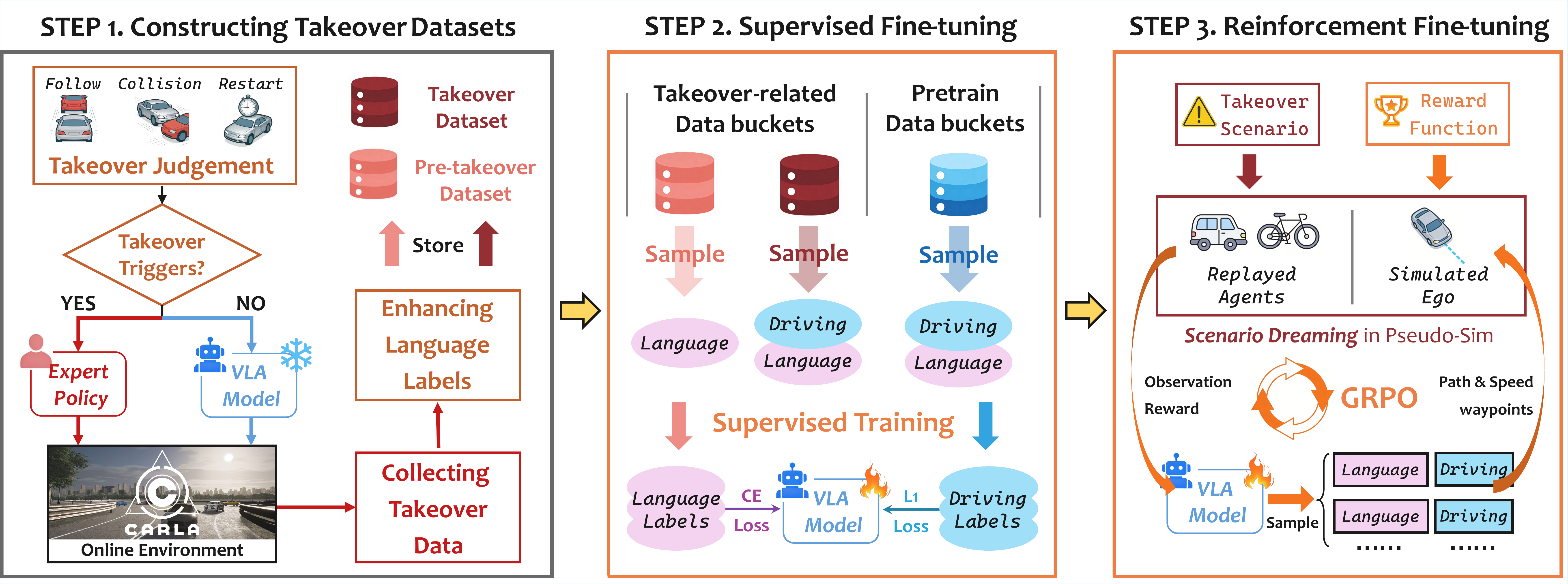}
  \caption{\textbf{Post-training pipiline for TakeVLA.}
    Starting from a pre-trained VLA model, each round consists of three key steps: 
    (1) online interaction with expert takeovers to construct pre-takeover and takeover datasets, 
    (2) supervised fine-tuning on the constructed datasets, 
    and (3) reinforcement fine-tuning via Scenario Dreaming in reconstructed takeover scenarios. 
    Multiple rounds progressively enhance language-conditioned driving performance.
  }
  \label{fig:framework}
\end{figure}

\section{Method}
\label{sec:method}

We propose \textbf{TakeVLA}, a post-training framework to enhance pre-trained VLA models for end-to-end autonomous driving. 
As illustrated in \cref{fig:framework}, TakeVLA starts from a pre-trained VLA and iteratively performs multiple rounds of takeover datasets construction and post-training to progressively improve performance. 
Each round comprises three key steps: 
(1) online interaction with expert takeovers to construct pre-takeover and takeover datasets, 
(2) supervised fine-tuning on the constructed datasets, 
and (3) reinforcement fine-tuning via Scenario Dreaming in reconstructed takeover scenarios.

 
\subsection{VLA Model Architecture and Pre-training}
\label{subsec:vla}

\begin{wrapfigure}{r}{0.45\textwidth}
    \centering
    \includegraphics[width=\linewidth]{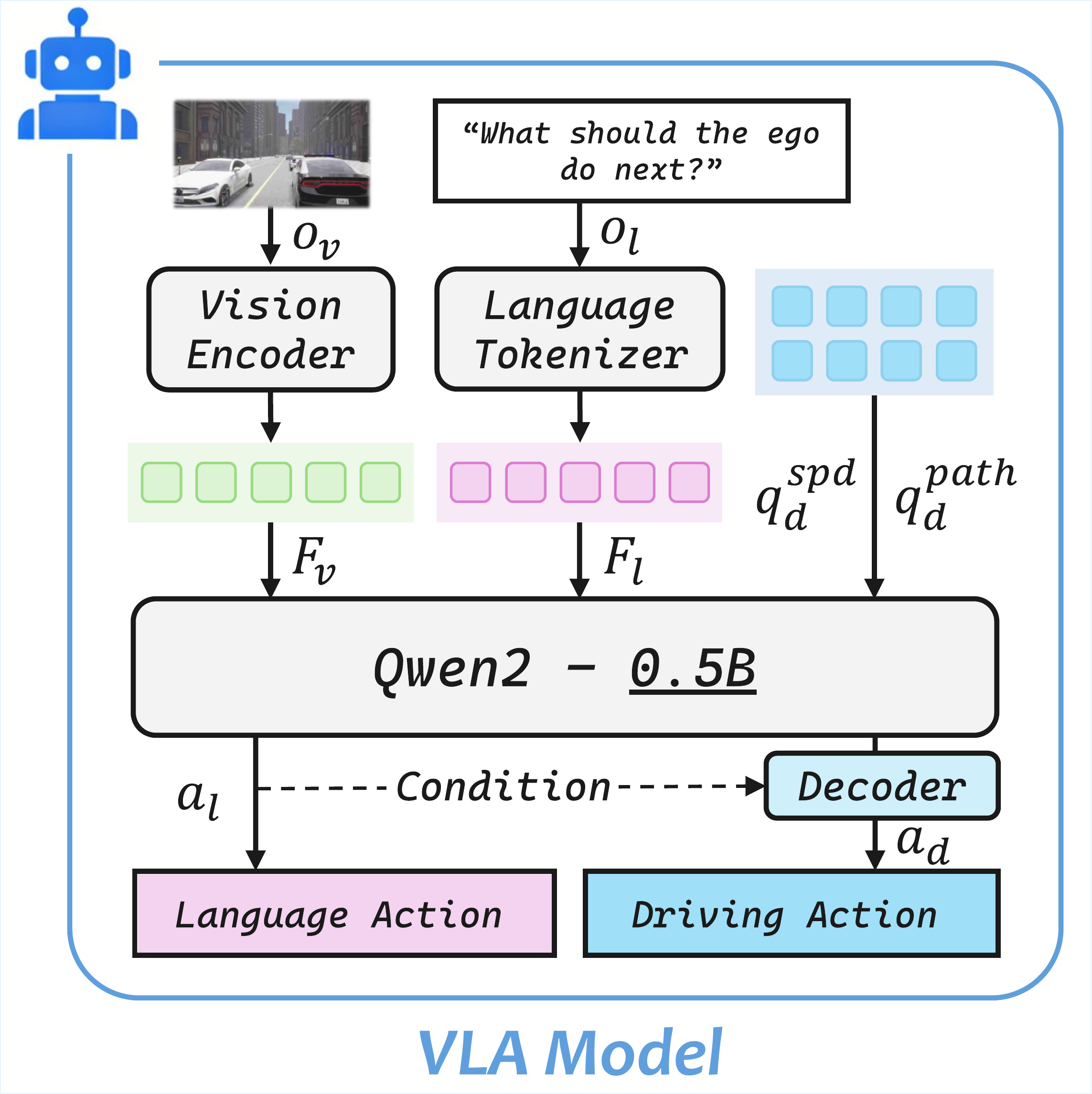}
    \caption{Architecture of the baseline VLA model.}
    \label{fig:vla}
\end{wrapfigure}

We adopt SimLingo~\cite{renz2025simlingo} as our baseline VLA model, which employs a Chain-of-Thought approach by conditioning driving actions on generated language. 
As illustrated in \cref{fig:vla}, the model takes a front-view camera image $o_v$ ($448 \times 896$) and a language instruction $o_l$ (including vehicle speed and task prompts) as inputs. 
$o_v$ is encoded into visual features $F_v$ via InternViT~\cite{chen2024internvl}, while $o_l$ is tokenized into language embeddings $F_l$. 
These features are fed into the LLM (Qwen2-0.5B~\cite{yang2024qwen2technicalreport}) to autoregressively generate the language action $a_l$. 
SimLingo decouples the driving action $a_d$ into path and speed waypoints, representing spatially and temporally equidistant coordinate sequences. 
Specifically, two learnable queries $q_d^{path}$ and $q_d^{spd}$ aggregate information from $F_l$, $F_v$, and $a_l$ in a single forward pass through the LLM.
The resulting query features are then passed to a Multi-Layer Perceptron decoder to produce the final coordinate sequences.

For simplicity, we omit the encoding of future navigation information in the figure and description. 
But note that they are also part of LLM's inputs, provided as additional language prompts or target point embeddings.

\textbf{Pre-training Process.}
Following SimLingo~\cite{renz2025simlingo}, we perform supervised pre-training on data collected by the privileged \textit{PDM-Lite}~\cite{sima2024drivelm, zimmerlin2024hidden} expert. 
The pre-training dataset $\mathcal{D}_{pretrain}$ contains approximately 3.1 million driving frames at 4 fps.
Pre-training uses the default configuration: bucket sampling for mitigating imbalanced data distribution, VQA tasks for vision-language understanding, and action dreaming tasks for robust language-action alignment.

\subsection{Constructing Takeover Datasets}

\textbf{Collecting Takeover Data.}
During online interaction between the VLA policy $\pi$ and the environment, the expert policy $\hat{\pi}$ runs in shadow mode to monitor and be ready to intervene~\cite{liu2025takead}. 
We define takeover judgment conditions $J = \{j_{Follow}, j_{Collision},$ $j_{Restart}\}$ to trigger takeover. 
Specifically, $j_{Follow}$ is triggered in car-following scenarios when the expert indicates deceleration, but the VLA fails to respond for 0.5 seconds, $j_{Collision}$ activates when a collision is predicted within 1 second under constant velocity, and $j_{Restart}$ engages when the ego-vehicle is stuck in high-density traffic due to missing lane-change opportunities.
Any triggered condition signals policy failure and prompts immediate expert takeover for $T_{take}$ frames.
By default, $T_{take}$ is set to 50 frames, corresponding to 2.5 seconds at 20 fps in the interaction environment.
For each triggered takeover at time $t$, we collect the raw takeover data sequence $\{o_l, o_v, \hat{a}_l,\hat{a}_d\}_t^{t+T_{take}}$, where $\hat{a}_d$ is the expert driving action, and $\hat{a}_l$ is the language label derived from the expert's internal logic.

\begin{figure}[tb]
  \centering
  \includegraphics[height=4.2cm]{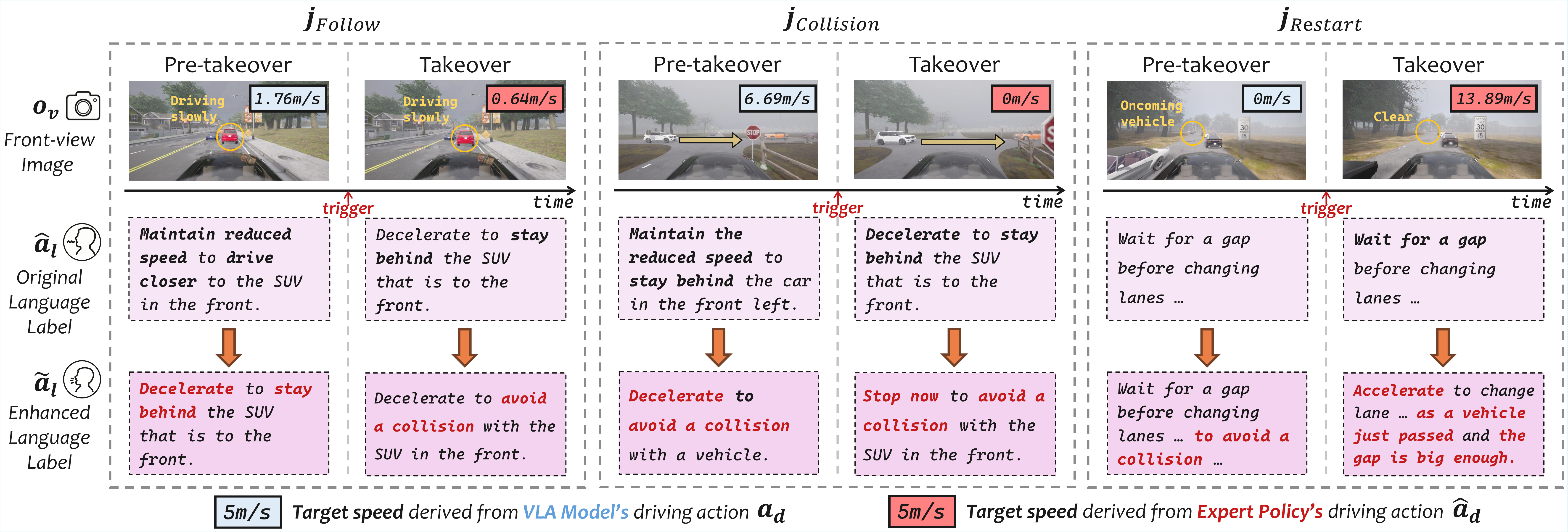}
  \caption{\textbf{Language label enhancement.}
    For each takeover trigger type (Follow, Collision, Restart), original language labels (middle) are refined (bottom) based on the specific cause. 
    Enhanced labels emphasize urgency and explicit causality, leading to more conservative and effective driving actions in critical scenarios.
  }
  \label{fig:enhance}
\end{figure}

\textbf{Enhancing Language Labels for Takeover Data.} 
The original language label $\hat{a}_l$ from the expert's internal logic may lack sufficient urgency or explicit causality, limiting the model's ability to learn timely avoidance in takeover scenarios. 
To address this, we enhance $\hat{a}_l$ based on the specific trigger $j \in J$ to produce the refined label $\tilde{a}_l$.
For example, when $j_{Collision}$ triggers emergency braking, the original label \textit{“Decelerate to stay behind ...”} becomes \textit{“Stop now to avoid a collision ...”}. 
\cref{fig:enhance} illustrates further examples.
The enhanced label $\tilde{a}_l$ forms the final takeover data sequence $ \{o_l, o_v, \tilde{a}_l,\hat{a}_d\}_t^{t+T_{take}}$.

\textbf{Pre-Takeover Data Construction.} 
To improve language output quality prior to the takeover moments and substantially enlarge safety margins, we additionally collect pre-takeover data.
Since no expert intervention occurs in this phase, no ground-truth expert driving actions are available. 
Thus, pre-takeover data sequence is represented as $\{ o_l, o_v, \tilde{a}_l\}_{t-T_{pre}}^{t-1}$, where $T_{pre}$ is the pre-takeover horizon.
The language labels $\tilde{a}_l$ are generated using the language label enhancement strategy described earlier based on the impending takeover reason $j \in J$,  yielding proactive warning-style output that enables earlier hazard detection.
By default, we set the pre-takeover horizon $T_{pre}=20$ frames, corresponding to 1 second at 20 fps in the online environment.

The resulting takeover and pre-takeover data sequences are stored in the corresponding datasets $\mathcal{D}_{takeover}$ and $\mathcal{D}_{pre-takeover} $ for subsequent post-training.

\subsection{Supervised Fine-tuning with Pre-takeover Data}

Using the takeover dataset $\mathcal{D}_{takeover}$ and pre-takeover dataset $\mathcal{D}_{pre-takeover}$, we first apply supervised fine-tuning (SFT) to the VLA model to simultaneously improve the quality of its language output and corresponding driving actions in takeover scenarios.
To mitigate catastrophic forgetting, we adopt a DAgger-style \cite{ross2011reduction} sampling strategy that mixes samples from the original pretraining dataset $\mathcal{D}_{pretrain}$ and the takeover datasets.
As described in \cref{subsec:vla}, the VLA model's supervised training employs bucket sampling. 
During SFT, we sample with a probability of $p$ from buckets built from $\mathcal{D}_{takeover}$ and $\mathcal{D}_{pre-takeover}$, and with a probability of $1-p$ from the original buckets of $\mathcal{D}_{pretrain}$.
By default, $p$ is set to 0.2 to prioritize takeover samples while preserving pre-training knowledge.

The sampled batches are supervised using cross-entropy loss on the language output $a_l$ against the enhanced language label $\tilde{a}_l$ and smooth-L1 loss on the predicted driving action $a_d$ against the expert driving action $\hat{a}_d$.
For pre-takeover samples that lack actual expert driving actions, we apply a mask that disables driving action supervision while retaining language supervision.
For a sampled instance $i$, the SFT loss is formulated as:
\begin{equation}
    \mathcal{L}^\text{SFT}_i = \mathcal{L}^\text{Cross-Entropy}_i(a_l, \tilde{a}_l) + (1 - m_i) \mathcal{L}^\text{Smooth-L1}_i(a_d, \hat{a}_d),
\end{equation}
where $m_i = 1$ for pre-takeover samples and $m_i = 0$ for takeover and pre-training samples.
Throughout the SFT process, we maintain the Action Dreaming task to preserve the VLA model's language-action alignment capability.

\subsection{Reinforcement Fine-tuning via Scenario Dreaming}

Following SFT, we employ reinforcement fine-tuning (RFT) to further optimize the VLA model for takeover scenarios. 
Since all RL training occurs in reconstructed scenes derived from samples of $\mathcal{D}_{takeover}$ and $\mathcal{D}_{pre-takeover}$, rather than real-world execution, we refer to this process as \textit{Scenario Dreaming}.

The RL exploration is conducted in a vector-space pseudo-simulation environment~\cite{Cao2025CORL, Dauner2024NEURIPS}, utilizing ground-truth annotations from the datasets. 
Specifically, for each sample, we simulate a 2-second future driving scene at 5 Hz, resulting in a simulation horizon of $H=10$. 
Ego-vehicle dynamics are modeled using a bicycle model, with control signals generated by the same PID controller used in online deployment. 
Other agents’ trajectories are replayed from recorded ground-truth future positions. 
The reward function is defined as:
\begin{equation}
    r_\tau = -r_\tau^{dis} - r_\tau^{col}, \quad R = \sum_{\tau=0}^{H} \gamma^\tau r_\tau
\end{equation}
where $\tau$ denotes the simulated time step, $r_\tau^{dis}$ is the L2 distance between the ego trajectory and the expert takeover trajectory at time $\tau$, and $r_\tau^{col} = 30$ indicates whether the ego vehicle’s bounding box overlaps with another agent’s box at time $\tau$, and 0 otherwise. 
$R$ is the discounted cumulative reward.

We use Group Relative Policy Optimization \cite{shao2024deepseekmath} (GRPO) for RL. 
For each takeover sample $i$, the old policy $\pi_{old}$ samples a group of candidate actions $\{a_j\}_{j=1}^G$ with a group size $G$ and evaluates the corresponding rewards $\{R_j\}_{j=1}^G$ via pseudo-simulation.
The current policy $\pi$ is then updated using the loss function:
\begin{equation}
    \mathcal{L}^{RFT}_i = \mathcal{L}^{GRPO}_i + \lambda \cdot \mathcal{L}^{KL}_i(\pi \| \pi_{ref}), \
    \mathcal{L}^{GRPO}_i = -\frac{1}{G} \sum_{j=1}^{G} A_j \cdot (\frac{\pi(a_j \mid o)}{\pi_{old}(a_j \mid o)}) 
\end{equation}
where $A_j = (R_{j}-\text{mean}(\{ R_{k} \}_{k=1}^{G})) /\ \text{std}(\{ R_{k} \}_{k=1}^{G})$ is the normalized group-relative advantage, and $\mathcal{L}_i^{KL}(\pi \| \pi_{ref})$ is the Kullback-Leibler (KL) divergence loss to prevent policy degradation due to excessive exploration, with $\pi_{ref}$ being the reference policy from the SFT stage and $\lambda$ being the KL loss weight.
We perform a single policy update per step without clipping to simplify the training as in~\cite{zhou2025autovla}.

\section{Experiments}
\label{sec:experiments}

\subsection{Experimental Setup}

\textbf{Benchmark and Metrics.}
We train and evaluate TakeVLA on the Bench2Drive (B2D) closed-loop benchmark~\cite{jia2024bench2drive}, which is built on the CARLA simulator~\cite{dosovitskiy2017carla}.
We follow the official Bench2Drive (B2D) protocol, which evaluates models on 220 challenging routes spanning 44 interactive scenario types, under diverse weather conditions and across different towns.
Multiple metrics are used to assess driving performance.
The \textbf{Driving Score (DS)} integrates Route Completion and Infraction Score, capturing both task accomplishment and adherence to traffic rules. 
The \textbf{Success Rate (SR)} represents the proportion of routes that are completed within the required time without any violations. 
\textbf{Efficiency} assesses the ego vehicle’s speed relative to surrounding traffic.
\textbf{Comfortness} measures the smoothness of the driving trajectory. 

\textbf{Takeover Dataset Collection and Usage.}
The post-training is performed iteratively over 3 rounds. 
As outlined in \cref{sec:method}, each round consists of takeover datasets construction, SFT, and RFT.
Specifically, the latest model is deployed in each round to collect approximately 40k takeover and 10k pre-takeover samples, totaling 50k takeover-related samples per round. 
To \textbf{prevent data leakage}, all data are collected exclusively from the CARLA Leaderboard v2 training routes, which are strictly disjoint from the evaluation routes. 
During SFT, we implement a bucket sampling strategy~\cite{renz2025simlingo} with 500k samples per epoch, where samples are drawn from the takeover-related and original pre-training datasets with probabilities $p$ and $1-p$, respectively. 
We set $p=0.2$ by default.
RFT further utilizes both pre-takeover and takeover data for Scenario Dreaming.

\textbf{Implementation Details.} 
For pre-training, we strictly follow the official SimLingo~\cite{renz2025simlingo} setup, training for 14 epochs with a total batch size of 64. 
For post-training, SFT uses the same batch size and runs for 8 epochs. 
RFT updates only the LLM while freezing other components to ensure training stability. 
We employ GRPO for policy optimization with a group size $G=8$, KL weight $\lambda=0.1$, and a sampling temperature of 1.0, training for 1 epoch with a total batch size of 32. 
The discount factor $\gamma=0.9$.
Both SFT and RFT use a learning rate of 3e-6 with a cosine annealing schedule. 
All experiments are conducted on 8$\times$ NVIDIA H20 GPUs. 
The LLM is updated via LoRA~\cite{hu2021loralowrankadaptationlarge}, maintaining consistent configurations across both pre-training and post-training stages.

\subsection{Main Results}

\begin{table}[tb]
  \caption{Closed-loop and Multi-ability performance of SoTA methods on the Bench2Drive leaderboard.
  }
  \label{tab:b2d}
  \centering
    \resizebox {\linewidth}{!}{
\begin{threeparttable}
\begin{tabular}{lc|>{\columncolor[gray]{0.9}}c>{\columncolor[gray]{0.9}}ccc|ccccc|>{\columncolor[gray]{0.9}}c}
\toprule
\multirow{2}{*}{\textbf{Methods}} & \multirow{2}{*}{\textbf{Reference}} & \multicolumn{4}{c|}{\textbf{Closed-loop Metric} $\uparrow$ } & \multicolumn{6}{c}{\textbf{Multi-ability (\%)} $\uparrow$} \\ 
\cmidrule{3-12} 
& & DS & SR (\%) & Efficiency & Comfortness & Merging & Overtaking & Emergency Brake & Give Way & Traffic Sign & \textbf{Mean} \\ 
\midrule
\rowcolor{blue!=8} \multicolumn{12}{c}{\textit{End-to-end Methods}} \\
UniAD~\cite{hu2023planning} & CVPR 23 & 45.81 & 16.36 & 129.21 & 43.58 & 12.16 & 20.00 & 23.64 & 10.00 & 13.89 & 15.89 \\
VAD~\cite{jiang2023vad} & ICCV 23 & 42.35 & 15.00 & 157.94 & 46.01 & 7.14 & 20.00 & 16.36 & 20.00 & 20.22 & 16.75 \\
SparseDrive~\cite{sun2025sparsedrive} & ICRA 25 & 44.54 & 16.71 & 170.21 & 48.63 & 12.18 & 23.19 & 17.91 & 20.00 & 20.98 & 17.45  \\
DriveTrans.-L.~\cite{jia2025drivetransformer} & ICLR 25 & 63.46 & 35.01 & 100.64 & 20.78 & 17.57 & 35.00 & 48.36 & 40.00 & 52.10 & 38.60 \\
DiffusionDrive~\cite{liao2025diffusiondrive} & CVPR 25 & 77.68 & 52.72 & 248.18 & 24.56 & 50.63 & 26.67 & 68.33 & 50.00 & 76.32 & 54.38 \\ 
Hydra-Next~\cite{li2025hydra} & ICCV 25 & 73.86 & 50.00 & 197.76 & 20.68 & 40.00 & 64.44 & 61.67 & 50.00 & 50.00 & 53.22 \\
TF$++$~\cite{zimmerlin2024hidden} & CVPRW 24 & 84.21 & 67.27 & - & - & 58.75 & 57.77 & 83.33 & 40.00 & 82.11 & 64.39 \\
Raw2Drive~\cite{yang2025raw2drive} & NeurIPS 25 & 71.36 & 50.24 & 214.17 & 22.42 & 43.35 & 51.11 & 60.00 & 50.00 & 62.26 & 53.34 \\
Hip-AD~\cite{tang2025hip} & ICCV 25 & 86.77 & 69.09 & 203.12 & 19.36 & 50.00 & 84.44 & 83.33 & 40.00 & 72.10 & 65.98 \\ 
TakeAD~\cite{liu2025takead} & RA-L 26 & 71.39 & 40.83 & 193.30 & 22.89 & 30.77 & 35.56 & 56.67 & 50.00 & 42.02 & 43.00 \\
\midrule
\rowcolor{blue!=8} \multicolumn{12}{c}{\textit{Vision Language Action Methods}} \\
DriveMOE\cite{yang2025drivemoe} & arXiv 25 & 74.22 & 48.64 & 175.96 & 15.31 & 34.67 & 40.00 & 65.45 & 40.00 & 59.44 & 47.91 \\ 
ORION\cite{fu2025orion} & ICCV 25 & 77.74 & 54.62 & 151.48 & 17.38 & 25.00 & 71.11 & 78.33 & 30.00 & 69.15 & 54.72 \\ 
MindDrive\cite{fu2025minddrive} & arXiv 26 & 78.04 & 55.09 & - & - & 32.89 & 75.56 & 68.33 & 50.00 & 57.89 & 56.94 \\ 
AutoVLA\cite{zhou2025autovla} & NeurIPS 25 & 78.84 & 57.73 & 146.93 & 39.33 & - & - & - & - & - & - \\ 
RecogDrive\cite{li2025recogdrive} & ICLR 26 & 71.36 & 45.45 & 138.18 & 17.45 & 29.73 & 20.00 & 69.09 & 20.00 & 71.34 & 42.03 \\ 
SimLingo$^*$~\cite{renz2025simlingo} & CVPR 25 & 84.79 & 64.84 & 248.79 & 28.61 & 51.90 & 53.33 & 85.00 & 50.00 & 78.42 & 63.73 \\ 
TakeVLA (\textbf{Ours}) & - & \textbf{89.72} & \textbf{73.73} & \textbf{249.01} & 30.27 & \textbf{63.64} & 64.44 & \textbf{91.67} & \textbf{50.00} & \textbf{85.48} & \textbf{71.05} \\ 
\midrule
\rowcolor{gray!=8} \multicolumn{12}{c}{\textit{Privileged Methods}} \\
\color{gray} \textit{Think2Drive}~\cite{li2024think2drive} & \color{gray} ECCV 24 & \color{gray} {91.85} & \color{gray} {85.41} & \color{gray} {269.14} & \color{gray} {25.97} & \color{gray} {81.27} & \color{gray} {83.92} & \color{gray} {90.24} & \color{gray} {90.00} & \color{gray} {87.67} & \color{gray} {86.26} \\ 
\color{gray} \textit{PDM-Lite$^*$}~\cite{sima2024drivelm} & \color{gray} CVPRW 24 & \color{gray} {96.25} & \color{gray} {83.49} & \color{gray} {213.47} & \color{gray} {22.33} & \color{gray} {82.05} & \color{gray} {91.11} & \color{gray} {81.67} & \color{gray} {80.00} & \color{gray} {70.97} & \color{gray} {81.16} \\ 
\bottomrule
\end{tabular}

\begin{tablenotes}
\footnotesize
\item[$*$] Reproduced using official implementation.
\end{tablenotes}

\end{threeparttable}}
\end{table}

\textbf{SoTA performance on Bench2Drive.}
\cref{tab:b2d} presents closed-loop driving performance of state-of-the-art (SoTA) methods on the B2D leaderboard. 
The left part shows overall driving metrics. 
Among all end-to-end methods, TakeVLA achieves the highest driving score and success rate, substantially outperforming our baseline method SimLingo~\cite{renz2025simlingo} by 4.93 points and 8.89\% , respectively. 
B2D further provides multi-ability evaluation, categorizing all test routes into different ability types, with each ability score defined as the success rate across all routes of that type. 
As shown in the right part of \cref{tab:b2d}, TakeVLA leads in the multi-ability mean score and nearly dominates all individual abilities.
These results demonstrate the effectiveness of our post-training framework in achieving superior closed-loop performance for driving VLA models.

\begin{table}[tb]
\caption{Closed-loop performance without rule-based post-process.}
\label{tab:no creep}
\centering
\setlength{\tabcolsep}{4pt}
\small
\resizebox{0.43\linewidth}{!}{
\begin{threeparttable}
\begin{tabular}{l|ccc}
\toprule
\multirow{2}{*}{\textbf{Methods}} & \multicolumn{3}{c}{\textbf{Closed-loop Metric}} \\
\cmidrule{2-4} 
 & DS $\uparrow$ & SR (\%) $\uparrow$ & TR (\%) $\downarrow$ \\
\midrule
SimLingo $^\dagger$ & 79.19 & 55.50 & 12.39 \\
TakeVLA (\textbf{Ours}) $^\dagger$ & \textbf{86.44} & \textbf{68.20} & \textbf{9.22} \\
\bottomrule
\end{tabular}

\begin{tablenotes}
\footnotesize
\item[$\dagger$] Without creep.
\end{tablenotes}

\end{threeparttable}}
\end{table}

\textbf{Improved Performance without Rule-based Post-process.}
Existing methods often rely on rule-based post-processing to boost performance metrics~\cite{sun2025sparsedrive, tang2025hip}. 
SimLingo uses the creep technique~\cite{chitta2023trans} to force forward motion when the ego-vehicle remains stationary for an extended period. 
While creep helps escape stuck states in practice, it introduces potential safety risks and highlights limitations in the model's intrinsic stop-go reasoning.
We address this by applying post-training with restart-triggered takeover data and language label enhancement.
As shown in \cref{tab:no creep}, the Timeout Rate (TR) metric measures the proportion of routes that fail due to timeout. 
Without any creep post-processing, TakeVLA maintains SoTA performance while achieving a noticeably higher driving score (+7.25) and lower timeout rate (-3.17\%) compared to SimLingo.
These results show that our post-training framework delivers strong closed-loop performance without rule-based post-processing, enhancing intrinsic stop-go reasoning.

\subsection{Ablation Studies}

In this section, we conduct ablation studies to evaluate the contribution of each component and key hyperparameter in TakeVLA.
All reported results, except for multi-round training, are obtained from \textbf{the first round} of post-training.
To directly quantify the model's proactive collision avoidance capability and safety margin, we introduce the average Time-to-Collision (TTC) metric, defined as:
\begin{equation}
    \text{TTC} = \frac{\sum_{i=1}^{N_a} \text{TTC}_i}{N_a}, \text{TTC}_i=\dfrac{d_i}{v_{\text{rel},i}},
\end{equation}
where $N_a$ is the total number of traffic participants that the ego encounters during evaluation.
The variable $v_{\text{rel}, i}$ denotes the relative velocity between the ego and the $i$-th participant, while $d_i$ represents their corresponding relative distance.
In practice, $\text{TTC}_i$ is computed only for participants located ahead of the ego, with positive relative velocity, and within 100m.

\begin{table}
  \caption{Ablation study on post-training techniques of TakeVLA.
  }
  \setlength{\tabcolsep}{4pt}
  \label{tab:ablation}
  \centering
    \resizebox {0.90\linewidth}{!}{
\begin{tabular}{c|cccc|ccc}
\toprule
\multirow{2}{*}{\textbf{ID}} & \multicolumn{4}{c|}{\textbf{Post-training Techniques}} & \multicolumn{3}{c}{\textbf{Closed-loop Metric} $\uparrow$} \\
\cmidrule{2-8}
& Takeover Data & Label Enhancement & Pre-takeover & Sceniario Dreaming & DS & SR (\%) & TTC (s) \\
\midrule
1 & \xmark & \xmark & \xmark & \xmark & 84.79 & 64.84 & 7.23 \\
\midrule
2 & \cmark & \xmark & \xmark & \xmark & 86.22 & 68.52 & 7.56 \\
3 & \cmark & \cmark & \xmark & \xmark & 86.73 & 68.04 & 7.53 \\
4 & \cmark & \cmark & \cmark & \xmark & 87.25 & 68.35 & 7.99 \\
5 & \cmark & \cmark & \cmark & \cmark & \cellcolor[gray]{0.9}87.79 & \cellcolor[gray]{0.9}68.81 & \cellcolor[gray]{0.9}8.16 \\
\bottomrule
\end{tabular}}
\end{table}

\textbf{Ablation on Post-training Techniques.}
\cref{tab:ablation} presents closed-loop performance under different combinations of post-training techniques. 
The baseline model (ID 1), without any post-training, achieves a driving score of 84.79, a success rate of 64.84\%, and an average TTC of 7.23 seconds.
Incorporating takeover data alone (ID 2) significantly improves the driving score by 1.43 points and TTC by 0.33 seconds, demonstrating that takeover data effectively mitigates distribution shift. 
Adding language label enhancement (ID 3) further boosts the driving score, confirming the benefit of refined language labels for better decision-making. 
Introducing pre-takeover supervision (ID 4) noticeably increases TTC by 0.43 seconds, validating its role in enlarging safety margins. 
Finally, enabling Scenario Dreaming (ID 5) yields the highest performance, with a driving score of 87.79 and a TTC of 8.16 seconds, showing that active exploration further optimizes the model in safety-critical scenarios.

\noindent
\begin{minipage}{0.44\textwidth}
   \centering
   \includegraphics[width=\linewidth]{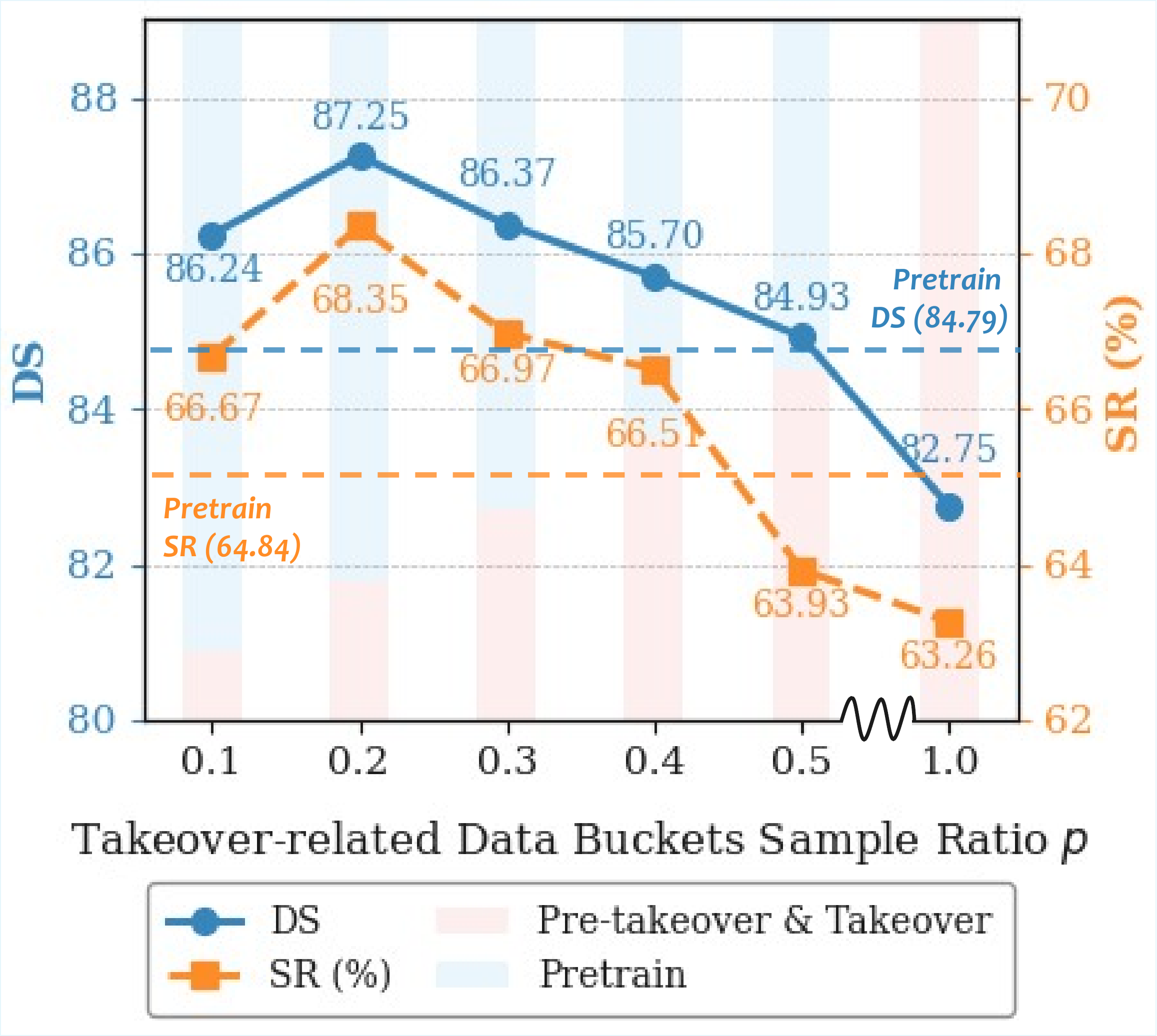}
   \captionof{figure}{\textbf{Impact of sampling ratio during SFT}.
   The colored bar visualizes the sampling ratio between pre-training and takeover-related buckets.} 
   \label{fig:sample_ratio}
\end{minipage}
\hfill
\begin{minipage}{0.48\textwidth}
   \centering
   \includegraphics[width=\linewidth]{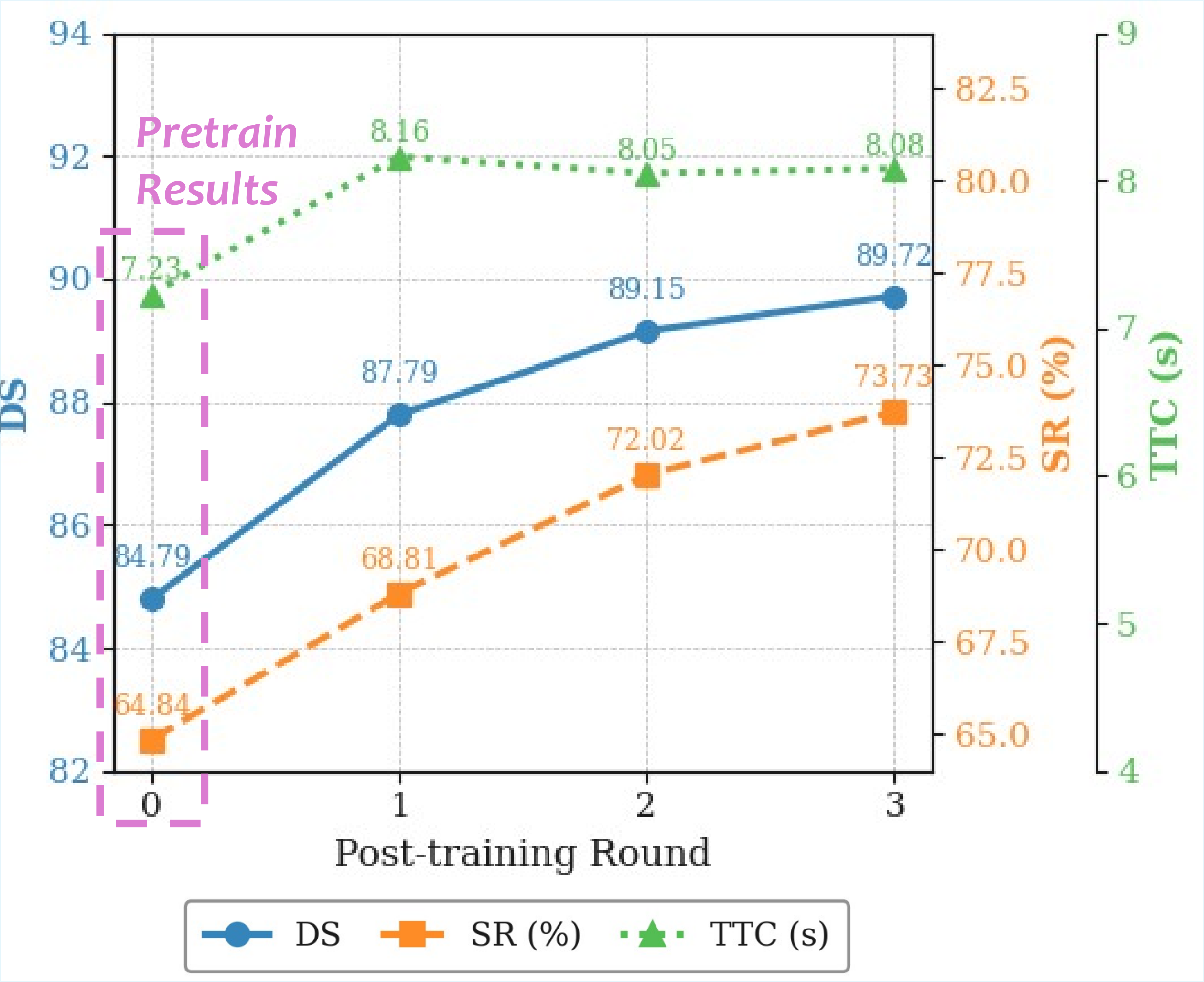}
   \captionof{figure}{\textbf{Multi-round post-training performance}.
   TakeVLA shows progressive performance gains in core driving metrics across rounds.} 
   \label{fig:multi_round}
\end{minipage}

\textbf{Ablation on Sample Ratio.} 
During SFT, we follow the bucket sampling strategy and assign sampling probability $p$ to buckets derived from takeover-related datasets $\mathcal{D}_{takeover}$ and $\mathcal{D}_{pre-takeover}$, and $1-p$ to buckets from the pre-training dataset $\mathcal{D}_{pretrain}$.
\cref{fig:sample_ratio} shows closed-loop performance as $p$ varies from 0.1 to 1.0, with the pre-training baseline as reference. 
Performance peaks at $p=0.2$, achieving the highest driving score of 87.25 and a success rate of 68.35\%, significantly outperforming the baseline. 
As $p$ increases beyond 0.2, both driving score and success rate gradually decline, reaching the lowest values at $p=1.0$.
This trend indicates that a moderate proportion of takeover samples effectively balances the learning of corrective behaviors while maintaining pre-training capabilities.
Consequently, we set $p=0.2$ as the default to achieve the best trade-off.

\begin{table}[tb]
  \caption{Effects of group size and reward design for GRPO training.
  }
  \setlength{\tabcolsep}{4pt}
  \label{tab:grpo}
  \centering
    \resizebox {0.52\linewidth}{!}{
\begin{tabular}{c|c|cc|ccc}
\toprule
\multirow{2}{*}{\textbf{ID}} & \textbf{Group} & \multicolumn{2}{c|}{\textbf{Reward terms}} & \multicolumn{3}{c}{\textbf{Closed-loop Metric} $\uparrow$} \\
\cmidrule{3-7}
& \textbf{size $G$} & Distance & Collision &  DS & SR (\%) & TTC (s) \\
\midrule
1 & 2 & \cmark & \cmark & 86.88 & 68.35 & 8.05 \\
2 & 4 & \cmark & \cmark & 87.33 & 67.28 & 8.01 \\
3 & 6 & \cmark & \cmark & 87.48 & 69.41 & 8.19 \\
4 & 8 & \cmark & \cmark & \cellcolor[gray]{0.9}87.79 & \cellcolor[gray]{0.9}68.81 & \cellcolor[gray]{0.9}8.16 \\
\midrule
5 & 8 & \cmark & \xmark & 87.22 & 68.81 & 7.54 \\
6 & 8 & \xmark & \cmark & 87.28 & 68.69 & 7.88 \\
\bottomrule
\end{tabular}}
\end{table}

\textbf{Ablation on GRPO Group Size and Reward Design.}
As shown in \cref{tab:grpo}, with both distance and collision terms enabled, performance steadily improves as the group size $G$ increases from 2 to 8 (IDs 1–4). 
The best overall results are achieved at $G=8$, indicating that a larger group size yields more stable group-relative advantage estimation and better policy optimization.
Ablating reward terms (IDs 5 and 6) reveals their complementary roles, as removing either term alone degrades performance, particularly in safety margins.
Specifically, removing the collision reward causes a clear drop in TTC, highlighting the importance of explicit collision avoidance optimization.
Removing the distance term leads to a milder TTC decline, indicating that collision rewards alone fail to provide comprehensive guidance. 
The distance term implicitly encodes certain hard-to-model reward signals from expert behaviors, thereby steering optimization toward better alignment with the demonstration policy.

\textbf{Ablation on the Number of Post-training Rounds.}
\cref{fig:multi_round} illustrates the progressive performance gains of TakeVLA across multiple post-training rounds.
Both driving score and success rate exhibit consistent improvements with each additional round, confirming that iterative post-training refines model behavior and enhances closed-loop performance. 
TTC reaches near-peak values after the first round and remains stable thereafter, with only minor fluctuations.
By the third round, TakeVLA achieves substantial improvements over the pre-trained baseline, increasing driving score by 4.93, success rate by 8.89\%, and TTC by 0.85 (a relative increase of 11.76\%).
The diminishing marginal gains, particularly TTC saturation after the first round, likely stem from information asymmetry between the expert and the VLA model~\cite{Nguyen2026CVPR}. 
The expert has access to comprehensive contextual information beyond the front-view camera input available to the VLA, which limits the model's ability to further close the performance gap in later rounds.
Future work will investigate expanded perceptual inputs and techniques from~\cite{Nguyen2026CVPR} to mitigate this asymmetry and enable more effective continual learning.

\subsection{Qualitative Results}
\begin{figure}[tb]
  \centering
  \includegraphics[height=8cm]{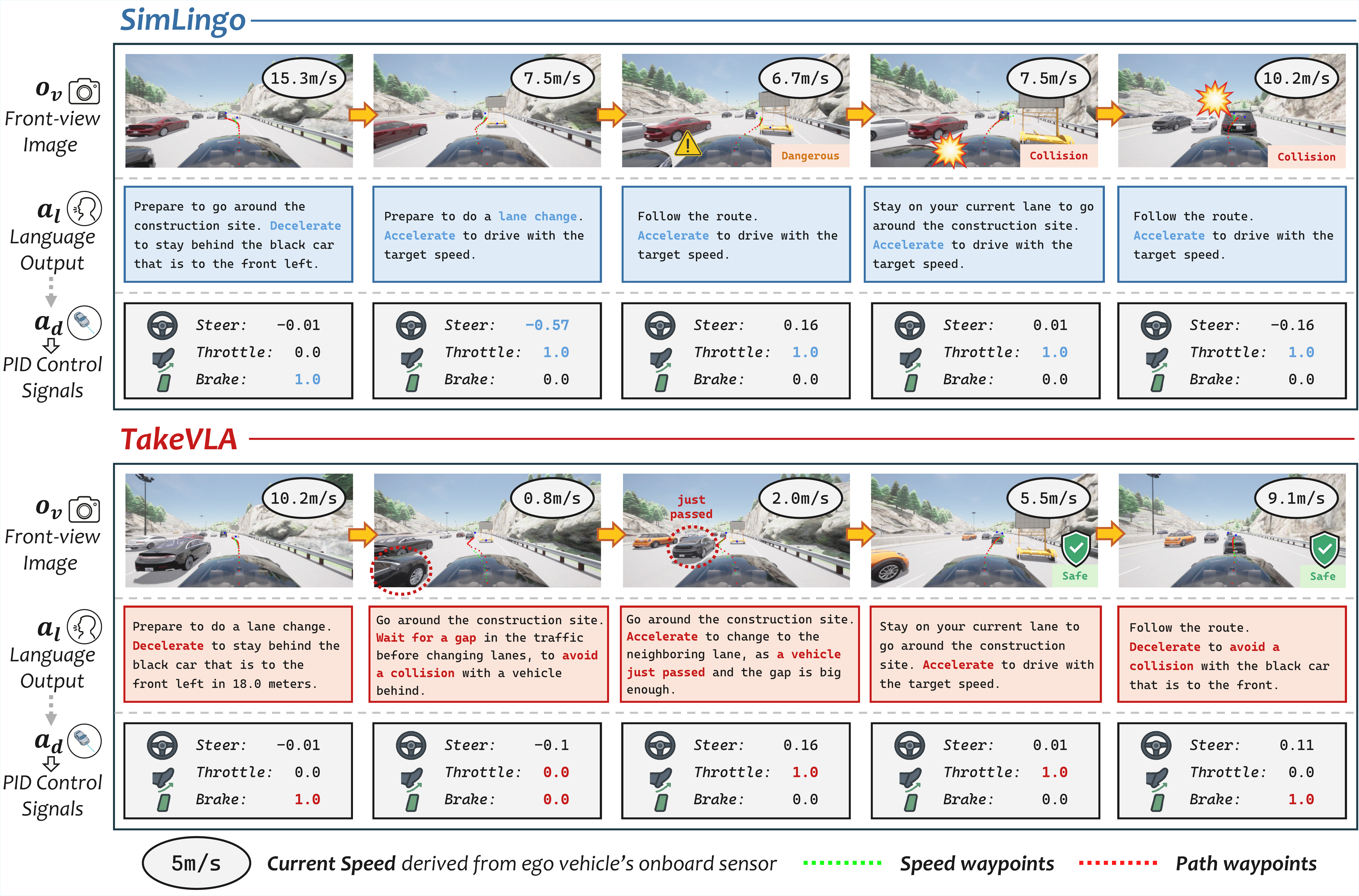}
  \caption{\textbf{Qualitative visualization of SimLingo and TakeVLA}.
  Each column shows a selected frame in temporal order.
  Driving action $a_d$ is drawn on the front-view image in the form of speed and path waypoints.
  The PID control signals are generated from $a_d$.
  We omit the language input $o_l$ for brevity.
  SimLingo generates insufficient language outputs, leading to unsafe lane changes and rear-end collisions.
  TakeVLA generates more anticipatory and context-aware language outputs, enabling safer gap selection, timely acceleration, and proactive deceleration to avoid collisions.
  }
  \label{fig:vis}
\end{figure}

\cref{fig:vis} provides a qualitative comparison of SimLingo and TakeVLA along a continuous driving route, which requires the ego-vehicle to bypass a construction site while yielding to continuous left-lane traffic.
The selected frames in sequence illustrate key moments of the route.
SimLingo generates brief language outputs that fail to anticipate lane-change risks, resulting in unsafe timing (\eg, initiating lane change without sufficient gap) and subsequent collisions, as well as delayed braking in car-following scenarios, causing rear-end collisions.
In contrast, TakeVLA produces richer, more context-aware language outputs that explicitly consider upcoming hazards and gaps, leading to successful collision avoidance during lane-changing and car-following while maintaining smooth driving. 
These examples illustrate how our proposed post-training framework enables safer and more reliable driving behavior.

\section{Conclusion}
In this paper, we presented TakeVLA, a novel post-training framework that effectively leverages expert takeover data to enhance pre-trained Vision-Language-Action (VLA) models for end-to-end autonomous driving.
By introducing pre-takeover language supervision tailored to takeover trigger conditions, TakeVLA enables proactive hazard anticipation through improved language output quality. 
Complementing this, our Scenario Dreaming mechanism performs reinforcement fine-tuning in reconstructed takeover scenarios, encouraging active exploration to achieve optimal policies beyond passive preference imitation.
Extensive closed-loop evaluation on the Bench2Drive benchmark demonstrates that TakeVLA achieves state-of-the-art performance while significantly improving safety margins.
We believe TakeVLA represents a practical step toward building safer, more reliable, and interpretable VLA systems for autonomous driving.

\bibliographystyle{splncs04}
\bibliography{main}

@String(CVPR  = {IEEE Conf. Comput. Vis. Pattern Recog.})

@String(NeurIPS = {Adv. Neural Inform. Process. Syst.})

@String(JMLR  = {J. Mach. Learn. Res.})

@String(CVPR  = {CVPR})

@String(NeurIPS = {NeurIPS})

@String(JMLR  = {JMLR})

@inproceedings{Shi-RSS-24, 
    AUTHOR    = {Lucy Xiaoyang Shi AND Zheyuan Hu AND Tony Z. Zhao AND Archit Sharma AND Karl Pertsch AND Jianlan Luo AND Sergey Levine AND Chelsea Finn}, 
    TITLE     = {{Yell At Your Robot: Improving On-the-Fly from Language Corrections}}, 
    BOOKTITLE = {Proceedings of Robotics: Science and Systems}, 
    YEAR      = {2024}, 
    ADDRESS   = {Delft, Netherlands}, 
    MONTH     = {July}, 
    DOI       = {10.15607/RSS.2024.XX.025} 
}

@inproceedings{hu2023planning,
  title={Planning-oriented autonomous driving},
  author={Hu, Yihan and Yang, Jiazhi and Chen, Li and Li, Keyu and Sima, Chonghao and Zhu, Xizhou and Chai, Siqi and Du, Senyao and Lin, Tianwei and Wang, Wenhai and others},
  booktitle={Proceedings of the IEEE/CVF conference on computer vision and pattern recognition},
  pages={17853--17862},
  year={2023}
}

@inproceedings{jiang2023vad,
  title={Vad: Vectorized scene representation for efficient autonomous driving},
  author={Jiang, Bo and Chen, Shaoyu and Xu, Qing and Liao, Bencheng and Chen, Jiajie and Zhou, Helong and Zhang, Qian and Liu, Wenyu and Huang, Chang and Wang, Xinggang},
  booktitle={Proceedings of the IEEE/CVF International Conference on Computer Vision},
  pages={8340--8350},
  year={2023}
}

@inproceedings{sun2025sparsedrive,
  title={Sparsedrive: End-to-end autonomous driving via sparse scene representation},
  author={Sun, Wenchao and Lin, Xuewu and Shi, Yining and Zhang, Chuang and Wu, Haoran and Zheng, Sifa},
  booktitle={2025 IEEE International Conference on Robotics and Automation (ICRA)},
  pages={8795--8801},
  year={2025},
  organization={IEEE}
}

@article{jia2025drivetransformer,
  title={Drivetransformer: Unified transformer for scalable end-to-end autonomous driving},
  author={Jia, Xiaosong and You, Junqi and Zhang, Zhiyuan and Yan, Junchi},
  journal={arXiv preprint arXiv:2503.07656},
  year={2025}
}

@inproceedings{liao2025diffusiondrive,
  title={Diffusiondrive: Truncated diffusion model for end-to-end autonomous driving},
  author={Liao, Bencheng and Chen, Shaoyu and Yin, Haoran and Jiang, Bo and Wang, Cheng and Yan, Sixu and Zhang, Xinbang and Li, Xiangyu and Zhang, Ying and Zhang, Qian and others},
  booktitle={Proceedings of the Computer Vision and Pattern Recognition Conference},
  pages={12037--12047},
  year={2025}
}

@inproceedings{li2025hydra,
  title={Hydra-next: Robust closed-loop driving with open-loop training},
  author={Li, Zhenxin and Wang, Shihao and Lan, Shiyi and Yu, Zhiding and Wu, Zuxuan and Alvarez, Jose M},
  booktitle={Proceedings of the IEEE/CVF International Conference on Computer Vision},
  pages={27305--27314},
  year={2025}
}

@article{zimmerlin2024hidden,
  title={Hidden biases of end-to-end driving datasets},
  author={Zimmerlin, Julian and Bei{\ss}wenger, Jens and Jaeger, Bernhard and Geiger, Andreas and Chitta, Kashyap},
  journal={arXiv preprint arXiv:2412.09602},
  year={2024}
}

@article{yang2025raw2drive,
  title={Raw2drive: Reinforcement learning with aligned world models for end-to-end autonomous driving (in carla v2)},
  author={Yang, Zhenjie and Jia, Xiaosong and Li, Qifeng and Yang, Xue and Yao, Maoqing and Yan, Junchi},
  journal={arXiv preprint arXiv:2505.16394},
  year={2025}
}

@inproceedings{tang2025hip,
  title={Hip-ad: Hierarchical and multi-granularity planning with deformable attention for autonomous driving in a single decoder},
  author={Tang, Yingqi and Xu, Zhuoran and Meng, Zhaotie and Cheng, Erkang},
  booktitle={Proceedings of the IEEE/CVF International Conference on Computer Vision},
  pages={25605--25615},
  year={2025}
}

@article{liu2025takead,
  title={TakeAD: Preference-Based Post-Optimization for End-to-End Autonomous Driving With Expert Takeover Data},
  author={Liu, Deqing and Gao, Yinfeng and Qian, Deheng and Zhang, Qichao and Ye, Xiaoqing and Han, Junyu and Zheng, Yupeng and Liu, Xueyi and Xia, Zhongpu and Ding, Dawei and others},
  journal={IEEE Robotics and Automation Letters},
  volume={11},
  number={2},
  pages={1738--1745},
  year={2025},
  publisher={IEEE}
}

@article{yang2025drivemoe,
  title={DriveMoE: Mixture-of-experts for vision-language-action model in end-to-end autonomous driving},
  author={Yang, Zhenjie and Chai, Yilin and Jia, Xiaosong and Li, Qifeng and Shao, Yuqian and Zhu, Xuekai and Su, Haisheng and Yan, Junchi},
  journal={arXiv preprint arXiv:2505.16278},
  year={2025}
}

@inproceedings{fu2025orion,
  title={Orion: A holistic end-to-end autonomous driving framework by vision-language instructed action generation},
  author={Fu, Haoyu and Zhang, Diankun and Zhao, Zongchuang and Cui, Jianfeng and Liang, Dingkang and Zhang, Chong and Zhang, Dingyuan and Xie, Hongwei and Wang, Bing and Bai, Xiang},
  booktitle={Proceedings of the IEEE/CVF International Conference on Computer Vision},
  pages={24823--24834},
  year={2025}
}

@article{fu2025minddrive,
  title={MindDrive: A Vision-Language-Action Model for Autonomous Driving via Online Reinforcement Learning},
  author={Fu, Haoyu and Zhang, Diankun and Zhao, Zongchuang and Cui, Jianfeng and Xie, Hongwei and Wang, Bing and Chen, Guang and Liang, Dingkang and Bai, Xiang},
  journal={arXiv preprint arXiv:2512.13636},
  year={2025}
}

@article{zhou2025autovla,
  title={Autovla: A vision-language-action model for end-to-end autonomous driving with adaptive reasoning and reinforcement fine-tuning},
  author={Zhou, Zewei and Cai, Tianhui and Zhao, Seth Z and Zhang, Yun and Huang, Zhiyu and Zhou, Bolei and Ma, Jiaqi},
  journal={arXiv preprint arXiv:2506.13757},
  year={2025}
}

@article{li2025recogdrive,
  title={Recogdrive: A reinforced cognitive framework for end-to-end autonomous driving},
  author={Li, Yongkang and Xiong, Kaixin and Guo, Xiangyu and Li, Fang and Yan, Sixu and Xu, Gangwei and Zhou, Lijun and Chen, Long and Sun, Haiyang and Wang, Bing and others},
  journal={arXiv preprint arXiv:2506.08052},
  year={2025}
}

@inproceedings{renz2025simlingo,
  title={Simlingo: Vision-only closed-loop autonomous driving with language-action alignment},
  author={Renz, Katrin and Chen, Long and Arani, Elahe and Sinavski, Oleg},
  booktitle={Proceedings of the Computer Vision and Pattern Recognition Conference},
  pages={11993--12003},
  year={2025}
}

@inproceedings{li2024think2drive,
  title={Think2drive: Efficient reinforcement learning by thinking with latent world model for autonomous driving (in carla-v2)},
  author={Li, Qifeng and Jia, Xiaosong and Wang, Shaobo and Yan, Junchi},
  booktitle={European conference on computer vision},
  pages={142--158},
  year={2024},
  organization={Springer}
}

@inproceedings{sima2024drivelm,
  title={Drivelm: Driving with graph visual question answering},
  author={Sima, Chonghao and Renz, Katrin and Chitta, Kashyap and Chen, Li and Zhang, Hanxue and Xie, Chengen and Bei{\ss}wenger, Jens and Luo, Ping and Geiger, Andreas and Li, Hongyang},
  booktitle={European conference on computer vision},
  pages={256--274},
  year={2024},
  organization={Springer}
}

@article{jia2024bench2drive,
  title={Bench2drive: Towards multi-ability benchmarking of closed-loop end-to-end autonomous driving},
  author={Jia, Xiaosong and Yang, Zhenjie and Li, Qifeng and Zhang, Zhiyuan and Yan, Junchi},
  journal={Advances in Neural Information Processing Systems},
  volume={37},
  pages={819--844},
  year={2024}
}

@article{chen2024vadv2,
  title={Vadv2: End-to-end vectorized autonomous driving via probabilistic planning},
  author={Chen, Shaoyu and Jiang, Bo and Gao, Hao and Liao, Bencheng and Xu, Qing and Zhang, Qian and Huang, Chang and Liu, Wenyu and Wang, Xinggang},
  journal={arXiv preprint arXiv:2402.13243},
  year={2024}
}

@inproceedings{zheng2025world4drive,
  title={World4drive: End-to-end autonomous driving via intention-aware physical latent world model},
  author={Zheng, Yupeng and Yang, Pengxuan and Xing, Zebin and Zhang, Qichao and Zheng, Yuhang and Gao, Yinfeng and Li, Pengfei and Zhang, Teng and Xia, Zhongpu and Jia, Peng and others},
  booktitle={Proceedings of the IEEE/CVF International Conference on Computer Vision},
  pages={28632--28642},
  year={2025}
}

@article{li2024enhancing,
  title={Enhancing end-to-end autonomous driving with latent world model},
  author={Li, Yingyan and Fan, Lue and He, Jiawei and Wang, Yuqi and Chen, Yuntao and Zhang, Zhaoxiang and Tan, Tieniu},
  journal={arXiv preprint arXiv:2406.08481},
  year={2024}
}

@inproceedings{shao2024lmdrive,
  title={Lmdrive: Closed-loop end-to-end driving with large language models},
  author={Shao, Hao and Hu, Yuxuan and Wang, Letian and Song, Guanglu and Waslander, Steven L and Liu, Yu and Li, Hongsheng},
  booktitle={Proceedings of the IEEE/CVF conference on computer vision and pattern recognition},
  pages={15120--15130},
  year={2024}
}

@article{wang2025alpamayo,
  title={Alpamayo-r1: Bridging reasoning and action prediction for generalizable autonomous driving in the long tail},
  author={Wang, Yan and Luo, Wenjie and Bai, Junjie and Cao, Yulong and Che, Tong and Chen, Ke and Chen, Yuxiao and Diamond, Jenna and Ding, Yifan and Ding, Wenhao and others},
  journal={arXiv preprint arXiv:2511.00088},
  year={2025}
}

@article{cai2025predictive,
  title={Predictive Preference Learning from Human Interventions},
  author={Cai, Haoyuan and Peng, Zhenghao and Zhou, Bolei},
  journal={arXiv preprint arXiv:2510.01545},
  year={2025}
}

@article{fang2025corevla,
  title={CoReVLA: A dual-stage end-to-end autonomous driving framework for long-tail scenarios via collect-and-refine},
  author={Fang, Shiyu and Cui, Yiming and Liang, Haoyang and Lv, Chen and Hang, Peng and Sun, Jian},
  journal={arXiv preprint arXiv:2509.15968},
  year={2025}
}

@book{sutton1998reinforcement,
  title={Reinforcement learning: An introduction},
  author={Sutton, Richard S and Barto, Andrew G and others},
  volume={1},
  number={1},
  year={1998},
  publisher={MIT press Cambridge}
}

@misc{yang2024qwen2technicalreport,
      title={Qwen2 Technical Report}, 
      author={An Yang and Baosong Yang and Binyuan Hui and Bo Zheng and Bowen Yu and Chang Zhou and Chengpeng Li and Chengyuan Li and Dayiheng Liu and Fei Huang and Guanting Dong and Haoran Wei and Huan Lin and Jialong Tang and Jialin Wang and Jian Yang and Jianhong Tu and Jianwei Zhang and Jianxin Ma and Jianxin Yang and Jin Xu and Jingren Zhou and Jinze Bai and Jinzheng He and Junyang Lin and Kai Dang and Keming Lu and Keqin Chen and Kexin Yang and Mei Li and Mingfeng Xue and Na Ni and Pei Zhang and Peng Wang and Ru Peng and Rui Men and Ruize Gao and Runji Lin and Shijie Wang and Shuai Bai and Sinan Tan and Tianhang Zhu and Tianhao Li and Tianyu Liu and Wenbin Ge and Xiaodong Deng and Xiaohuan Zhou and Xingzhang Ren and Xinyu Zhang and Xipin Wei and Xuancheng Ren and Xuejing Liu and Yang Fan and Yang Yao and Yichang Zhang and Yu Wan and Yunfei Chu and Yuqiong Liu and Zeyu Cui and Zhenru Zhang and Zhifang Guo and Zhihao Fan},
      year={2024},
      eprint={2407.10671},
      archivePrefix={arXiv},
      primaryClass={cs.CL},
      url={https://arxiv.org/abs/2407.10671}, 
}

@inproceedings{ross2011reduction,
  title={A reduction of imitation learning and structured prediction to no-regret online learning},
  author={Ross, St{\'e}phane and Gordon, Geoffrey and Bagnell, Drew},
  booktitle={Proceedings of the fourteenth international conference on artificial intelligence and statistics},
  pages={627--635},
  year={2011},
  organization={JMLR Workshop and Conference Proceedings}
}

@article{shao2024deepseekmath,
  title={Deepseekmath: Pushing the limits of mathematical reasoning in open language models},
  author={Shao, Zhihong and Wang, Peiyi and Zhu, Qihao and Xu, Runxin and Song, Junxiao and Bi, Xiao and Zhang, Haowei and Zhang, Mingchuan and Li, YK and Wu, Yang and others},
  journal={arXiv preprint arXiv:2402.03300},
  year={2024}
}

@misc{hu2021loralowrankadaptationlarge,
      title={LoRA: Low-Rank Adaptation of Large Language Models}, 
      author={Edward J. Hu and Yelong Shen and Phillip Wallis and Zeyuan Allen-Zhu and Yuanzhi Li and Shean Wang and Lu Wang and Weizhu Chen},
      year={2021},
      eprint={2106.09685},
      archivePrefix={arXiv},
      primaryClass={cs.CL},
      url={https://arxiv.org/abs/2106.09685}, 
}

@article{chitta2023trans,
  author={Chitta, Kashyap and Prakash, Aditya and Jaeger, Bernhard and Yu, Zehao and Renz, Katrin and Geiger, Andreas},
  journal={IEEE Transactions on Pattern Analysis and Machine Intelligence}, 
  title={TransFuser: Imitation With Transformer-Based Sensor Fusion for Autonomous Driving}, 
  year={2023},
  volume={45},
  number={11},
  pages={12878-12895},
  doi={10.1109/TPAMI.2022.3200245}}

@inproceedings{Nguyen2026CVPR,
	author = {Long Nguyen and Micha Fauth and Bernhard Jaeger and Daniel Dauner and Maximilian Igl and Andreas Geiger and Kashyap Chitta},
	title = {LEAD: Minimizing Learner-Expert Asymmetry in End-to-End Driving},
	booktitle = {Conference on Computer Vision and Pattern Recognition (CVPR)},
	year = {2026},
}

@inproceedings{Cao2025CORL, 
	author = {Wei Cao and Marcel Hallgarten and Tianyu Li and Daniel Dauner and Xunjiang Gu and Caojun Wang and Yakov Miron and Marco Aiello and Hongyang Li and Igor Gilitschenski and Boris Ivanovic and Marco Pavone and Andreas Geiger and Kashyap Chitta}, 
	title = {Pseudo-Simulation for Autonomous Driving}, 
	booktitle = {Conference on Robot Learning (CoRL)}, 
	year = {2025}, 
}

@inproceedings{Dauner2024NEURIPS,
	title = {NAVSIM: Data-Driven Non-Reactive Autonomous Vehicle Simulation and Benchmarking},
	author = {Daniel Dauner and Marcel Hallgarten and Tianyu Li and Xinshuo Weng and Zhiyu Huang and Zetong Yang and Hongyang Li and Igor Gilitschenski and Boris Ivanovic and Marco Pavone and Andreas Geiger and Kashyap Chitta},
	booktitle = {Advances in Neural Information Processing Systems (NeurIPS)},
	year = {2024},
}

@ARTICLE{zheng2026plan,
  author={Zheng, Yupeng and Xing, Zebin and Zhang, Qichao and Jin, Bu and Li, Pengfei and Zheng, Yuhang and Xia, Zhongpu and Chen, Yaran and Zhao, Dongbin},
  journal={IEEE Transactions on Cognitive and Developmental Systems}, 
  title={PlanAgent: A Multi-modal Large Language Agent for Closed-loop Vehicle Motion Planning}, 
  year={2026},
  volume={},
  number={},
  pages={1-14},
  doi={10.1109/TCDS.2026.3664120}}

@INPROCEEDINGS{BlackK-RSS-25, 
    AUTHOR    = {Kevin Black AND Noah Brown AND Danny Driess AND Adnan Esmail AND Michael Robert Equi AND Chelsea Finn AND Niccolo Fusai AND Lachy Groom AND Karol Hausman AND Brian Ichter AND Szymon Jakubczak AND Tim Jones AND Liyiming Ke AND Sergey Levine AND Adrian Li-Bell AND Mohith Mothukuri AND Suraj Nair AND Karl Pertsch AND Lucy Xiaoyang Shi AND Laura Smith AND James Tanner AND Quan Vuong AND Anna Walling AND Haohuan Wang AND Ury Zhilinsky}, 
    TITLE     = {{{$\pi$0}: A Vision-Language-Action Flow Model for General Robot Control}}, 
    BOOKTITLE = {Proceedings of Robotics: Science and Systems}, 
    YEAR      = {2025}, 
    ADDRESS   = {LosAngeles, CA, USA}, 
    MONTH     = {June}, 
    DOI       = {10.15607/RSS.2025.XXI.010} 
}

@inproceedings{
kim2024openvla,
title={Open{VLA}: An Open-Source Vision-Language-Action Model},
author={Moo Jin Kim and Karl Pertsch and Siddharth Karamcheti and Ted Xiao and Ashwin Balakrishna and Suraj Nair and Rafael Rafailov and Ethan P Foster and Pannag R Sanketi and Quan Vuong and Thomas Kollar and Benjamin Burchfiel and Russ Tedrake and Dorsa Sadigh and Sergey Levine and Percy Liang and Chelsea Finn},
booktitle={8th Annual Conference on Robot Learning},
year={2024},
url={https://openreview.net/forum?id=ZMnD6QZAE6}
}

@ARTICLE{zhang2022trajgen,
  author={Zhang, Qichao and Gao, Yinfeng and Zhang, Yikang and Guo, Youtian and Ding, Dawei and Wang, Yunpeng and Sun, Peng and Zhao, Dongbin},
  journal={IEEE Transactions on Intelligent Transportation Systems}, 
  title={TrajGen: Generating Realistic and Diverse Trajectories With Reactive and Feasible Agent Behaviors for Autonomous Driving}, 
  year={2022},
  volume={23},
  number={12},
  pages={24474-24487},
  doi={10.1109/TITS.2022.3202185}}

@ARTICLE{gao2024piwm,
  author={Gao, Yinfeng and Zhang, Qichao and Ding, Da-Wei and Zhao, Dongbin},
  journal={IEEE Transactions on Intelligent Vehicles}, 
  title={Dream to Drive With Predictive Individual World Model}, 
  year={2024},
  volume={9},
  number={12},
  pages={8224-8238},
  doi={10.1109/TIV.2024.3408830}}

@misc{tian2025simscale,
      title={SimScale: Learning to Drive via Real-World Simulation at Scale}, 
      author={Haochen Tian and Tianyu Li and Haochen Liu and Jiazhi Yang and Yihang Qiu and Guang Li and Junli Wang and Yinfeng Gao and Zhang Zhang and Liang Wang and Hangjun Ye and Tieniu Tan and Long Chen and Hongyang Li},
      year={2025},
      eprint={2511.23369},
      archivePrefix={arXiv},
      primaryClass={cs.CV},
      url={https://arxiv.org/abs/2511.23369}, 
}

@INPROCEEDINGS{gao2019compare,
  author={Gao, Yinfeng and Liu, Yuqi and Zhang, Qichao and Wang, Yu and Zhao, Dongbin and Ding, Dawei and Pang, Zhonghua and Zhang, Yueming},
  booktitle={2019 Tenth International Conference on Intelligent Control and Information Processing (ICICIP)}, 
  title={Comparison of Control Methods Based on Imitation Learning for Autonomous Driving}, 
  year={2019},
  volume={},
  number={},
  pages={274-281},
  doi={10.1109/ICICIP47338.2019.9012185}}

@misc{wang2026meanfuser,
      title={MeanFuser: Fast One-Step Multi-Modal Trajectory Generation and Adaptive Reconstruction via MeanFlow for End-to-End Autonomous Driving}, 
      author={Junli Wang and Xueyi Liu and Yinan Zheng and Zebing Xing and Pengfei Li and Guang Li and Kun Ma and Guang Chen and Hangjun Ye and Zhongpu Xia and Long Chen and Qichao Zhang},
      year={2026},
      eprint={2602.20060},
      archivePrefix={arXiv},
      primaryClass={cs.CV},
      url={https://arxiv.org/abs/2602.20060}, 
}

@misc{rafailov2024dpo,
      title={Direct Preference Optimization: Your Language Model is Secretly a Reward Model}, 
      author={Rafael Rafailov and Archit Sharma and Eric Mitchell and Stefano Ermon and Christopher D. Manning and Chelsea Finn},
      year={2024},
      eprint={2305.18290},
      archivePrefix={arXiv},
      primaryClass={cs.LG},
      url={https://arxiv.org/abs/2305.18290}, 
}

@article{ouyang2022training,
  title={Training language models to follow instructions with human feedback},
  author={Ouyang, Long and Wu, Jeffrey and Jiang, Xu and Almeida, Diogo and Wainwright, Carroll and Mishkin, Pamela and Zhang, Chong and Agarwal, Sandhini and Slama, Katarina and Ray, Alex and others},
  journal={Advances in neural information processing systems},
  volume={35},
  pages={27730--27744},
  year={2022}
}

@inproceedings{peng2024improving,
  title={Improving agent behaviors with rl fine-tuning for autonomous driving},
  author={Peng, Zhenghao and Luo, Wenjie and Lu, Yiren and Shen, Tianyi and Gulino, Cole and Seff, Ari and Fu, Justin},
  booktitle={European Conference on Computer Vision},
  pages={165--181},
  year={2024},
  organization={Springer}
}

@inproceedings{zheng2024genad,
  title={Genad: Generative end-to-end autonomous driving},
  author={Zheng, Wenzhao and Song, Ruiqi and Guo, Xianda and Zhang, Chenming and Chen, Long},
  booktitle={European Conference on Computer Vision},
  pages={87--104},
  year={2024},
  organization={Springer}
}

@inproceedings{chen2024internvl,
  title={Internvl: Scaling up vision foundation models and aligning for generic visual-linguistic tasks},
  author={Chen, Zhe and Wu, Jiannan and Wang, Wenhai and Su, Weijie and Chen, Guo and Xing, Sen and Zhong, Muyan and Zhang, Qinglong and Zhu, Xizhou and Lu, Lewei and others},
  booktitle={Proceedings of the IEEE/CVF conference on computer vision and pattern recognition},
  pages={24185--24198},
  year={2024}
}

@article{jiang2025alphadrive,
  title={Alphadrive: Unleashing the power of vlms in autonomous driving via reinforcement learning and reasoning},
  author={Jiang, Bo and Chen, Shaoyu and Zhang, Qian and Liu, Wenyu and Wang, Xinggang},
  journal={arXiv preprint arXiv:2503.07608},
  year={2025}
}

@article{zeng2025futuresightdrive,
  title={Futuresightdrive: Thinking visually with spatio-temporal cot for autonomous driving},
  author={Zeng, Shuang and Chang, Xinyuan and Xie, Mengwei and Liu, Xinran and Bai, Yifan and Pan, Zheng and Xu, Mu and Wei, Xing and Guo, Ning},
  journal={arXiv preprint arXiv:2505.17685},
  year={2025}
}

@article{chen2024end,
  title={End-to-end autonomous driving: Challenges and frontiers},
  author={Chen, Li and Wu, Penghao and Chitta, Kashyap and Jaeger, Bernhard and Geiger, Andreas and Li, Hongyang},
  journal={IEEE Transactions on Pattern Analysis and Machine Intelligence},
  volume={46},
  number={12},
  pages={10164--10183},
  year={2024},
  publisher={IEEE}
}

@article{rowe2025poutine,
  title={Poutine: Vision-language-trajectory pre-training and reinforcement learning post-training enable robust end-to-end autonomous driving},
  author={Rowe, Luke and de Schaetzen, Rodrigue and Girgis, Roger and Pal, Christopher and Paull, Liam},
  journal={arXiv preprint arXiv:2506.11234},
  year={2025}
}

@article{yang2025worldrft,
  title={WorldRFT: Latent World Model Planning with Reinforcement Fine-Tuning for Autonomous Driving},
  author={Yang, Pengxuan and Lu, Ben and Xia, Zhongpu and Han, Chao and Gao, Yinfeng and Zhang, Teng and Zhan, Kun and Lang, XianPeng and Zheng, Yupeng and Zhang, Qichao},
  journal={arXiv preprint arXiv:2512.19133},
  year={2025}
}

@inproceedings{dosovitskiy2017carla,
  title={CARLA: An open urban driving simulator},
  author={Dosovitskiy, Alexey and Ros, German and Codevilla, Felipe and Lopez, Antonio and Koltun, Vladlen},
  booktitle={Conference on robot learning},
  pages={1--16},
  year={2017},
  organization={PMLR}
}
\end{document}